\documentclass[conference]{IEEEtran}
\IEEEoverridecommandlockouts
\usepackage{cite}
\usepackage{amsmath,amssymb,amsfonts}
\usepackage{algorithmic}
\usepackage{graphicx}
\usepackage{textcomp}
\usepackage{xcolor}
\usepackage{comment}
\usepackage{hyperref}

\def\BibTeX{{\rm B\kern-.05em{\sc i\kern-.025em b}\kern-.08em
    T\kern-.1667em\lower.7ex\hbox{E}\kern-.125emX}}

\begin{document}

\title{Integrating Large Language Models for UAV Control in Simulated Environments: A Modular Interaction Approach\\

\thanks{}
}

\author{
\begin{tabular}{cc}
  Abhishek Phadke* & Alihan Hadimlioglu \\
  School of Engineering and Computing & Department of Computer Science \\
  Christopher Newport University & Texas A\&M University - Corpus Christi \\
  Newport News, Virginia & Corpus Christi, Texas \\
  abhishek.phadke@tamucc.edu & alihan.hadimlioglu@tamucc.edu \\
  *Corresponding author & \\ 
  & \\ % Add an empty row for spacing
  Tianxing Chu & Chandra N Sekharan \\
  Department of Computer Science  & Department of Computer Science \\
  Conrad Blucher Institute for Surveying and Science & Texas A\&M University - Corpus Christi \\
  Texas A\&M University - Corpus Christi & Corpus Christi, Texas \\
  Corpus Christi, Texas  & chandra.sekharan@tamucc.edu \\
  tianxing.chu@tamucc.edu \\
\end{tabular}

}
\maketitle

\begin{abstract}
The intersection of LLMs (Large Language Models) and UAV (Unoccupied Aerial Vehicles) technology represents a promising field of research with the potential to enhance UAV capabilities significantly. This study explores the application of LLMs in UAV control, focusing on the opportunities for integrating advanced natural language processing into autonomous aerial systems. By enabling UAVs to interpret and respond to natural language commands, LLMs simplify the UAV control and usage, making them accessible to a broader user base and facilitating more intuitive human-machine interactions. The paper discusses several key areas where LLMs can impact UAV technology, including autonomous decision-making, dynamic mission planning, enhanced situational awareness, and improved safety protocols. Through a comprehensive review of current developments and potential future directions, this study aims to highlight how LLMs can transform UAV operations, making them more adaptable, responsive, and efficient in complex environments. A template development framework for integrating LLMs in UAV control is also described. Proof of Concept results that integrate existing LLM models and popular robotic simulation platforms are demonstrated. The findings suggest that while there are substantial technical and ethical challenges to address, integrating LLMs into UAV control holds promising implications for advancing autonomous aerial systems.
\end{abstract}


\section{Introduction}
LLMs (Large Language Models) are advanced artificial intelligence systems designed to understand, generate, and interact with human language at a large scale. These models are trained on vast amounts of data, enabling them to grasp the complexities of language, including grammar, context, and nuances. LLMs can perform various natural language processing tasks, such as translation, summarization, question answering, and conversational interaction. They drive many modern AI (Artificial Intelligence) applications, including chatbots, virtual assistants, and content-generation tools. The development of LLMs has significantly advanced the field of AI, making it possible for machines to engage with humans in more natural and meaningful ways.

UAVs (Unoccupied Aerial Vehicles) are at the forefront of robotics research, having found use in many civil applications \cite{Phadke2023ExaminingAS} such as land surveying, agriculture \cite{Radoglou2020ACO}, product delivery, emergency communications \cite{Kang2023LyapunovOB} as well as military applications of airspace defense, reconnaissance \cite{Wei2022MultiUAVsCR} and military support \cite{Kerr2020BattlefieldMB}.

\begin{figure*}[htbp]
\centering
\includegraphics[width=\textwidth]{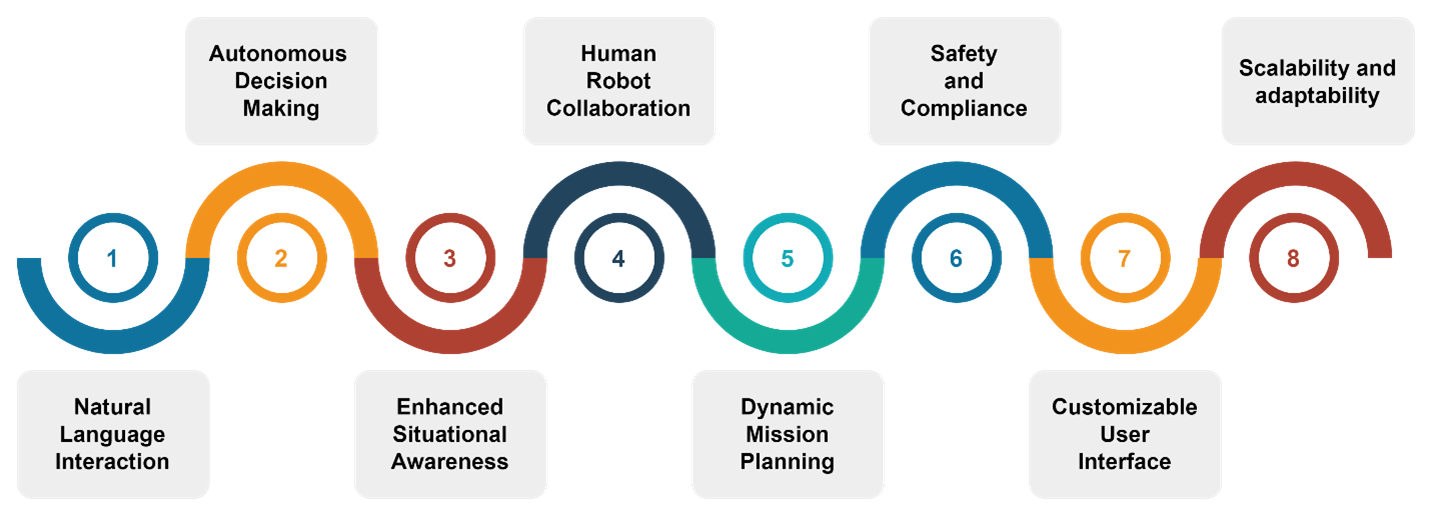}
\caption{Possibilities and opportunities for LLM integration in UAV control}
\label{fig1:llm_integration}
\end{figure*}

Integrating LLMs into UAV control presents a transformative opportunity in autonomous systems. By harnessing the natural language processing capabilities of LLMs, UAVs can be controlled and instructed using conversational language, making the interaction more intuitive and accessible. This integration allows for the interpretation of complex commands and the execution of intricate maneuvers, enhancing the UAVs' adaptability and responsiveness in dynamic environments. For instance, a UAV with an LLM could understand a command like "Survey the area for wildlife" and autonomously plan and execute the necessary flight path. This fusion of LLMs with UAV technology simplifies the human-machine interface and opens up new possibilities for deploying UAVs in various applications, ranging from disaster response to environmental monitoring, where real-time decision-making and flexibility are crucial.

\section{Exploring opportunities for LLM integration with UAVs}
This section highlights the various possibilities of integrating LLMs with UAV control, frameworks, and simulations to harness their decision-making capabilities, code generation, and response.

The key opportunities highlighted in Figure \ref{fig1:llm_integration} are discussed below.

LLMs can transform operators' interactions with UAVs by enabling control through natural language commands. This approach simplifies the user interface, making it more accessible to a wider audience, including those without technical expertise. For instance, an operator could instruct a UAV to "Survey the coastline for signs of pollution," and the UAV would then autonomously interpret and execute the command. This includes asking the user for follow-up prompts such as the starting point of the coastline survey, flight altitude, and the expected data collection that is required from the survey. This could potentially reduce handling time and pilot learning curves. Since LLMs are an evolving field, any improvement could also improve functions that are lower in the hierarchy. For example, with modern LLMs learning to interact with the user via audio input prompts, the user may be able to converse with the UAV agent in real-time through an audio channel.

As the UAVs navigate complex environments, they face various dynamic conditions and disruptions that can impede performance. Integrating systemic resilience is a challenging concept in UAV control that can be significantly reduced with automated decision-making. By leveraging LLMs' contextual understanding and predictive capabilities, aerial vehicles can make informed decisions to adapt to an evolving situation graciously.

The ability of LLMs to process and analyze vast amounts of textual and image-based data from various sources, such as weather reports or geographical information, can provide UAVs with enhanced situational awareness and improve their operational effectiveness. For example, a specially trained LLM can harness data from onboard or ground-based sensors to examine wind gusts and image data from the UAV sensors. 

This can then be analyzed in real time to adjust vital operational parameters such as flying height, fuel calculations, and sensor adjustments. 

LLMs can improve coordination in collaborative tasks by facilitating more intuitive communication between humans and UAVs. For example, in a search and rescue operation, a UAV could understand and execute complex instructions from a human team leader, such as scanning particular coordinates of the ROI (Region Of Interest) map for the missing person or target or making multiple low-level passes of a region in a specific grid or pattern.

UAVs with LLMs can dynamically update their mission plans based on real-time information and natural language instructions, allowing for more flexible and responsive operations. As mentioned above, strong wind gusts, increased presence of obstacles, or the requirement to continuously transmit large data payloads may proportionally change fuel consumption metrics. An LLM-backed control interface may be able to examine trends in consumption and accurately predict vital maneuvers such as returning to base or deploying replacement agents.

\begin{table*}[htbp]
\caption{A Comprehensive Review of Recent LLM-based Innovations in Autonomous Robotics}
\begin{center}
            \begin{tabular}{| p{0.15\linewidth} | p{0.04\linewidth}|p{0.74\linewidth}|}
            \hline
             \textbf{Study Reference} & \textbf{Year} & \textbf{Method Description} \\ [0.5ex] 
             \hline
             \cite{Liu2023LLMBasedHC} & 2023 & Advanced LLM-based logical interface for autonomous robotic manipulation  \\ 
             \hline
             \cite{Lykov2023LLMBRAInAI} & 2023 & LLM-based generation of robot behavior trees  \\
             \hline
             \cite{Palnitkar2023ChatSimUS} & 2023 & An LLM-supported natural language platform to directly control the simulation environment in which autonomous vehicles work  \\
             \hline
             \cite{Piggott2023NetGPTA} & 2023 & An offensive LLM supported Chatbot that uses UAVs for attack applications \\
             \hline
             \cite{Zhao2023ChatWT}  & 2023 & An interactive LLM-supported perception framework to instruct actions and reason over the results\\              
             \hline
             \cite{Chen2024RoboScriptCG}  & 2024 & A code generation pipeline and benchmark for LLM-generated code for robotic manipulation \\ 
             \hline
             \cite{Latif20243PLLMPP}  & 2023 & Probabilistic real-time path planning and feedback using popular OpenAI LLM and a simulated environment\\         
             \hline
            \cite{Mahadevan2024GenerativeER}  & 2023 & An approach to generate and modify expressive robot motion using LLM\\ 
             \hline
             \cite{Mu2024RoboCodeXMC}  & 2023 & A Large Vision Language model interface for LLM and robotic control integration\\ 
             \hline
             \cite{Nwankwo2024TheCI}  & 2023 & Harnessing LLMs and multimodal vision-language models to enable human interactions with autonomous robots\\ 
             \hline
             \cite{Wang2024ConformalTL}  & 2023 & A hierarchical LLM-supported planner to design and assign multiple high-level subtasks to be executed by mobile robots in a temporal and logical order\\ 
             [1ex] 
             \hline
            \end{tabular}
            \label{table1}
            \end{center}
    \end{table*}    

While the above information focuses more on the benefits of LLM integration in a UAV regarding capability, the sections below address additional benefits of integration, the first starting with safety and compliance. By interpreting relevant geospatial information, images, and guidelines, LLMs can help UAVs understand and comply with regulatory requirements, safety protocols, and airspace restrictions. This capability is crucial for ensuring that UAVs operate safely and legally, minimizing the risk of accidents and violations. Additionally, agent control through user interfaces can be suitably scaled so that inexperienced users have less difficulty adapting and using aerial systems. The inherent nature of LLMs to be continuously updated and trained on new data will allow UAV systems to evolve and adapt to changing requirements and environments over time. Adaption to changing conditions is one of the primary methods of resilience integration in dynamic engineered systems \cite{Woods2015FourCF}.

\begin{figure}[htbp]
  \centering
  \includegraphics[width=3.4in]{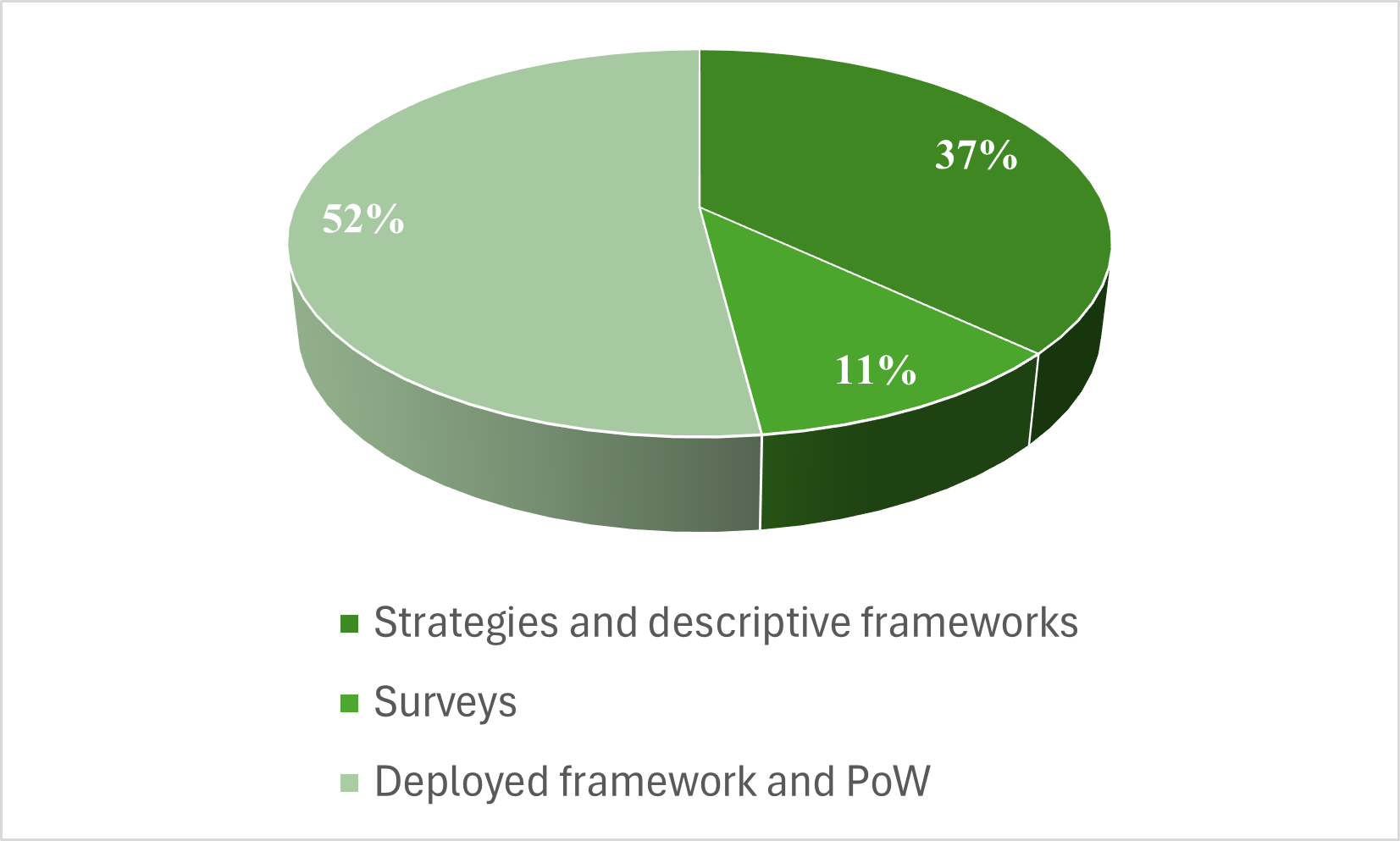}
  \caption{Visualization of the spread of examined research in three categories}
  \label{fig2}
\end{figure}

\section{Summary of current research trends}

LLMs and their integration with robotics have surged in the past three years. The advancements in UAV control and the different modules of UAV development that can act as viable areas of integration have been discussed above. The following section briefly discusses current implementations of LLMs in controlling unoccupied and industrial robotics. In this context, the search is not just limited to aerial vehicles. Instead, any recent study in the broad domain of robotic interactions based on LLMs is included. The recent work is shown in Table \ref{table1}. This was done to examine the different methods used and explore the possibility of using them for further integration in the UAV control framework. Considering the novelty of the topic, the article search for the survey and descriptive study spanned from 2023 to August 2024. Research items were generally focused on three major areas. The first is "Strategies and descriptive frameworks for LLM use with robotics," the second is "Surveys on current and future possibilities," and finally, "Deployed frameworks that demonstrate actual applications and usage of LLMs with robotics." The general spread of the examined research is visualized in Figure \ref{fig2}.

\section{Development of the framework template}

\begin{figure}[htbp]
  \centering
  \includegraphics[width=3.5in]{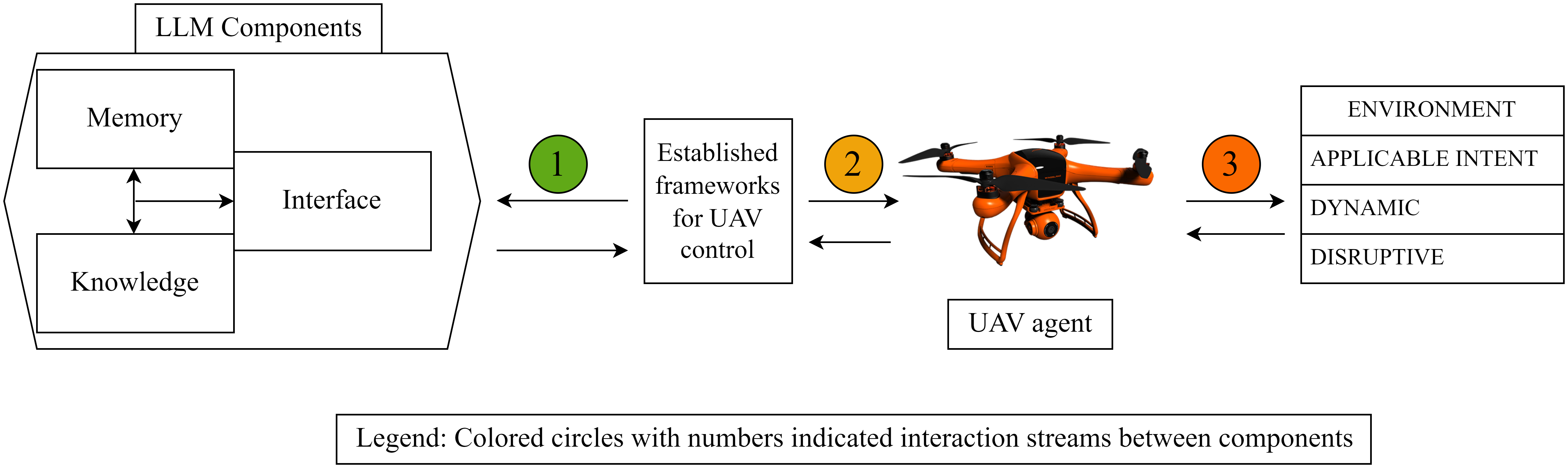}
  \caption{Boilerplate framework for interaction streams and LLM communication with UAV agent and environment}
  \label{fig3}
\end{figure}

Based on insights gained from this study and the survey conducted regarding the current cross-integration of UAVs and LLMs, a boilerplate development framework is described in this section. Figure \ref{fig3} shows the framework template developed and used in this proof of concept.

The framework is essentially divided into four primary areas. The LLM model, which functions through an API access key, generates the code commands required for the agent to execute actions in the environment. The established frameworks for GNC (Guidance Navigation Control) govern the agent's actions and facilitate the bidirectional collection, production, and transfer of action and reaction information in the framework. The simulated UAV agent and the environment reside within a simulation platform that can render these components graphically and produce the data required by the GNC framework and LLM to execute further decisions. This description primarily applies to simulated environments. In real-world experiments, the UAV agent and the environment are physical components. The figure also depicts colored circles with numbers 1,2 and 3. They are the interaction streams between different components of the framework. Interaction streams refer to the processes that occur between two components due to the cascaded module outputs coming from any direction in a linear methodology. For example, the interaction stream of the order 1, 2, and 3 is used to program and examine initial commands given by the LLM framework.

Similarly, as the UAV reacts in the environment and produces actions like motion, obstacle sensing, collision, or data collection, these actions go back to the framework and through it to the LLM to generate the next cycle of decisions. In this case, the order of the interaction stream is 3, 2, and 1. The parameters of the different framework components used in this study are summarized in Table \ref{table2}.

\begin{table}[htbp]
\caption{Details the primary framework components, specific platforms used, and their listed function}
\begin{center}
            \begin{tabular}{| p{0.20\linewidth} | p{0.24\linewidth}|p{0.35\linewidth}|}
            \hline
             \textbf{Framework Component} & \textbf{Specific Model / Platform Used} & \textbf{Primary Function} \\ [0.5ex] 
             \hline
             LLM & ChatGPT models (various) by OpenAI \cite{OpenAiChatGPT} & This is the primary LLM used for decision, control, and code generation actions that filter down the proposed framework and produce results in CEP-2.  \\ 
             \hline
             GNC/CEP-1 (Code Execution Platform) & MATLAB by Mathworks  & Acts as the host platform for the LLM/API wrapper.\\
             \hline
             LLM/API user interface  & MATGPT \cite{Takeuchi2024MatGPT} & Third-party developed (MIT license) API using a wrapper to use the LLM model in conjunction with CEP-1. \\
             \hline
             CEP-2/Agent and environment simulation platform & CoppeliaSim \cite{Rohmer2013CoppeliaSimFA} & Generates the simulation effectives such as the UAV agent, obstacles, and the 3D environment.  \\
             [1ex] 
             \hline  
            \end{tabular}
            \label{table2}
            \end{center}
\end{table}

\begin{figure}[htbp]
\centering
\includegraphics[width=3.4in]{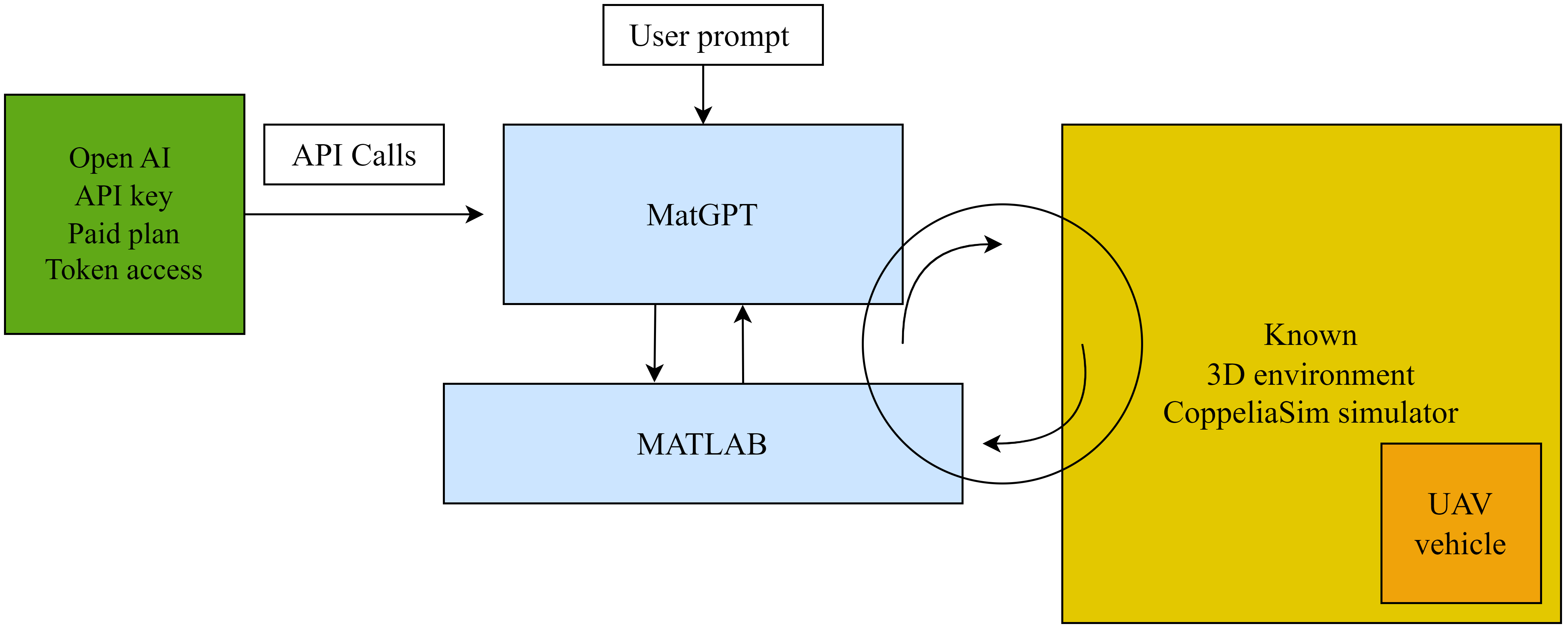}
\caption{Visualization of the different components and their connections as outlined in the boilerplate framework}
\label{fig4} 
\end{figure}

Figure \ref{fig4} shows the hierarchy for the experiment and simulation development. The chosen components interact with each other to produce the conversion from the original prompt to pre-coded functions in CEP-1, which are then transferred to CEP-2 for the simulation of the agent and the environment.

\section{Experiment methodology and interaction stream structure}

\begin{figure}[htbp]
\centering
\includegraphics[width=3.4in]{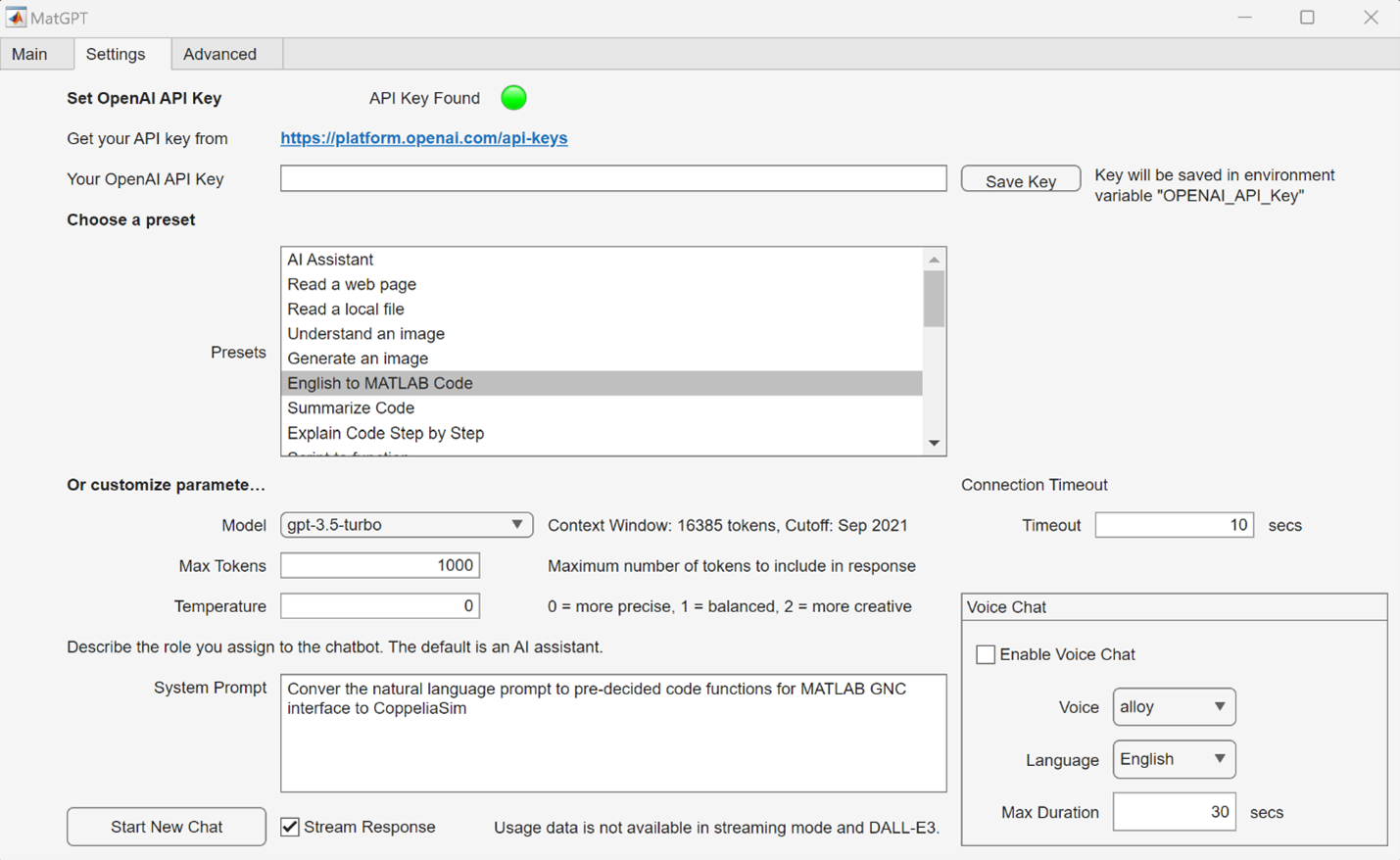}
\caption{MatGPT screen \cite{Takeuchi2024MatGPT}}
\label{fig5} 
\end{figure}

To initially access the LLM model via API and transfer code to MATLAB, we utilize MatGPT \cite{Takeuchi2024MatGPT}, a third-party wrapper with an MIT license. Figure \ref{fig5} displays the main MatGPT screen, where critical LLM parameters are defined, such as the specific model, maximum token allowance,  LLM temperature (balancing predictability and creativity in generated content), and system prompts. Developed by Toshiaki Takeuchi, this MATLAB application generates the order of predefined MATLAB code functions.The application uses the LLMs with MATLAB code \cite{LLMswithMATLAB} to connect MATLAB® to OpenAI API.

The following assumptions define the initial proof of concept environment.

\begin{itemize}
  \item There is a single, static obstacle in the path of the UAV
  \item The entire environment is pre-mapped on a local coordinate frame, and the UAV agent can reference its position with respect to the environment
  \item The obstacle position and dimensions are known, and this information is a part of the pre-mapped environment
These limitations were imposed to keep the complexity of the interaction stream structure low during initial experiments. Additionally, since the LLM access via the API user interface running on the CEP-1 wrapper happens through a paid, per-API call model, this also helped keep costs low.
\end{itemize} 

\begin{figure}[htbp]
\centering
\includegraphics[width=3.4in]{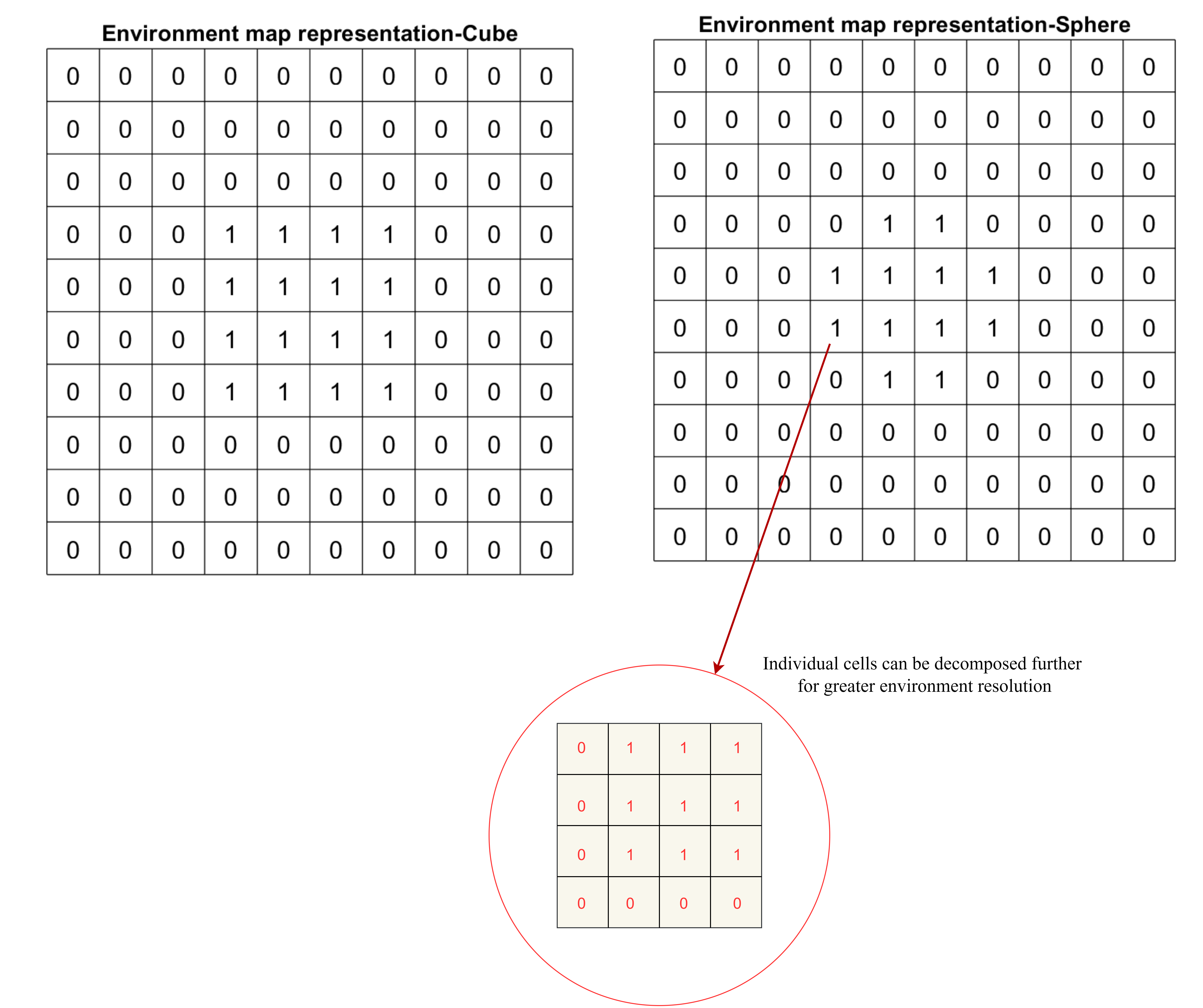}
\caption{Top-down visualization of the environment with obstacle information. Cells can be further decomposed, as shown in the red circle for finer control}
\label{fig6} 
\end{figure}

\begin{figure}[htbp]
\centering
\includegraphics[width=3in]{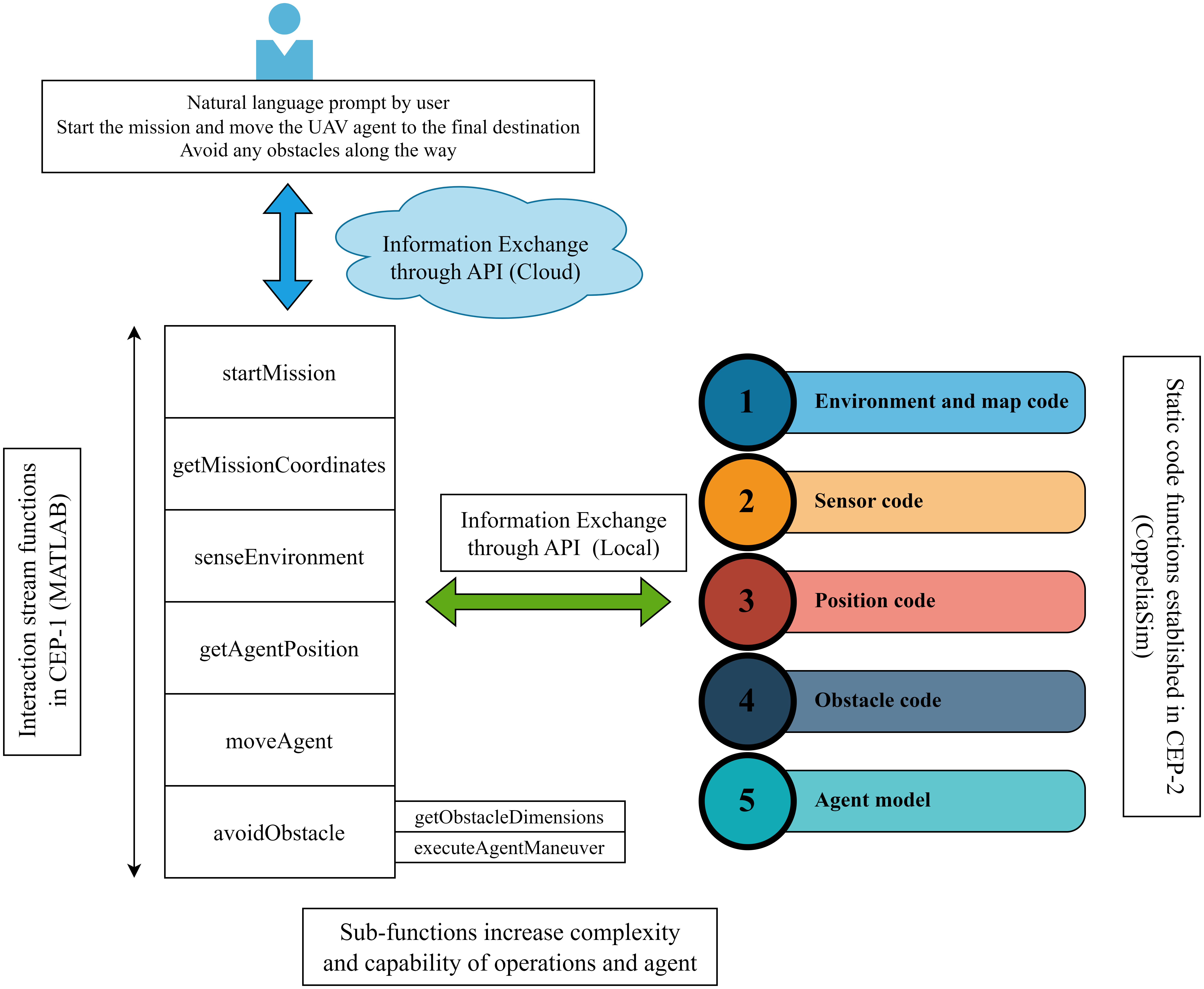}
\caption{Complete methodology setup showing interaction stream functions, API locations, and CEP 1 and 2 placements}
\label{fig7} 
\end{figure}

\begin{table}[htbp]
\caption{Summary of Interaction streams and their descriptions}
            \begin{center}
            \begin{tabular}{| p{0.30\linewidth} | p{0.60\linewidth}|}
            \hline
             \textbf{Interaction Stream} & \textbf{Function} \\ [0.5ex] 
             \hline
             startMission & This starts the mission and calls the pre-defined interaction stream functions in the CEP-1.\\ 
             \hline
             getMissionCoordinates & This checks the start and end mission coordinates for the UAV agent.\\
             \hline
             senseEnvironment  & This loads the pre-mapped 3D environment as an XYZ heightmap to sense environment dynamics.\\
             \hline
             getAgentPosition & This uses the superimposed local coordinate system on the XYZ environment heightmap shown in Figure \ref{fig6}  to generate the Agent position. 1s indicate presence of obstacle and 0s indicate no obstacle\\
             \hline
             moveAgent  & This uses the agent model and control code in CEP-2 to adjust actuators and produce roll, pitch, and yaw motions.\\
             \hline
             avoidObstacle\newline
             getObstacleDimensions\newline
             executeAgentManeuver 
  & This compares the obstacle dimensions from the XYZ heightmap and agent position to generate an avoidance path around obstacle dimensions.

The executeAgentManeuver is a sub-call that asks for ½ or 3-second iteration control commands in the x, y, and z plane for the agent to continue motion in the suggested direction.
\\
             [1ex] 
             \hline  
            \end{tabular}
            \label{table3}
            \end{center}
\end{table}

Mission parameters were mapped in the GNC CEP-1 platform before interaction via remote API to CEP-2 was initiated. This isolated problems between platform interactions and the underlying code before testing the complete pathway. The mapped results of the initial missions are also shown in the next section.

\begin{table}[htbp]
\caption{Description of three basic mission parameters used in the experiment}
\begin{center}
            \begin{tabular}{| p{0.25\linewidth} | p{0.65\linewidth}|}
            \hline
             \textbf{Mission Number} & \textbf{Mission Parameters} \\ [0.5ex] 
             \hline
             Mission 1 & The agent moves from start to destination. It encounters a cube-shaped obstacle. The agent bypasses it using a turn-based movement.\\ 
             \hline
             Mission 2 & The agent moves from start to destination. It encounters a cube-shaped obstacle. The agent bypasses it using an altitude change movement.\\
             \hline
             Mission 3 & The agent moves from start to destination. It encounters a spherical obstacle. The agent bypasses it using a turn-based movement after circumventing the obstacle.\\
             [1ex] 
             \hline  
            \end{tabular}
            \label{table4}
            \end{center}
\end{table}

\section{Results and Discussion}
\begin{figure}[htbp]
\centering
\includegraphics[width=3.3in]{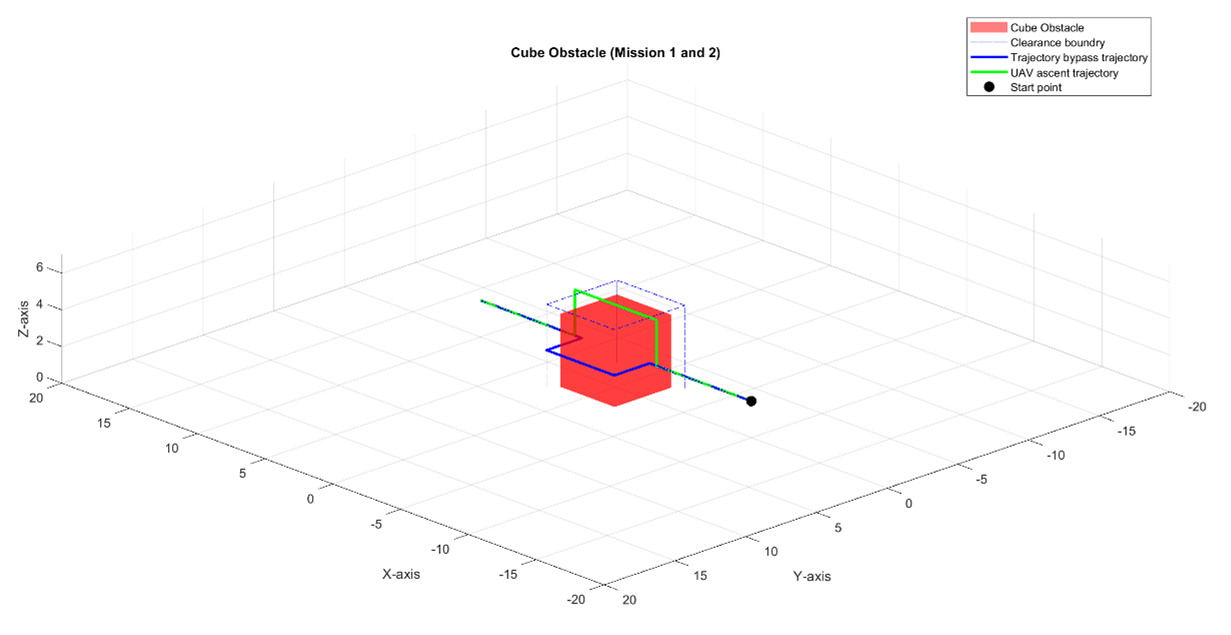}
\caption{Mission 1 and 2 executions in CEP-1 environment}
\label{fig8} 
\end{figure}

\begin{figure}[htbp]
\centering
\includegraphics[width=3.3in]{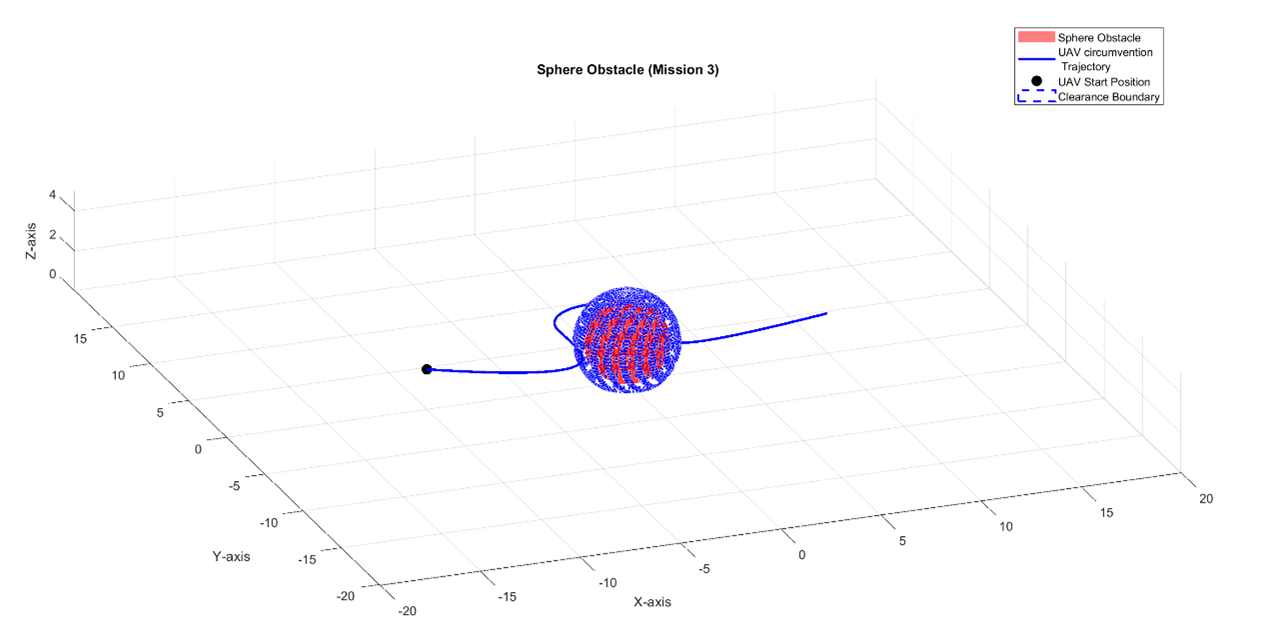}
\caption{Mission 3 execution in CEP-1 environment}
\label{fig9} 
\end{figure}

\begin{figure}[htbp]
\centering
\includegraphics[width=3.2in]{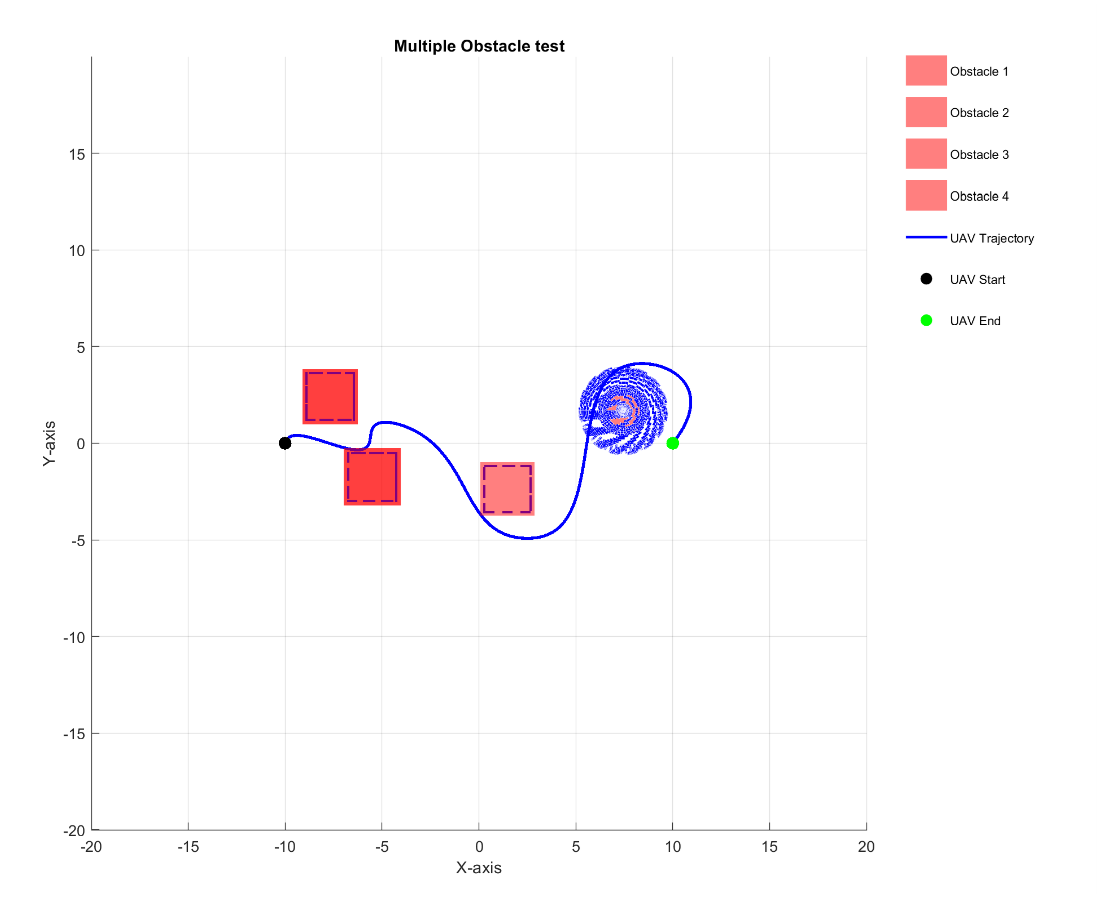}
\caption{complex environment test conducted in CEP-1 as a feasibility test for future work}
\label{fig10} 
\end{figure}

Initial missions carried out using direct control of the UAV agent through the interaction streams being called manually are shown in Figure \ref{fig8}, Figure \ref{fig9}, and Figure \ref{fig10}.

In missions 1 and 2, the obstacle is a single cube. The UAV uses the same set of interaction streams; however, in mission 2, an additional constraint is placed in the prompt instructing the UAV to circumvent any obstacle by conducting an altitude change only. This effectively blocks the agent from bypassing the agent using the known map information. Instead, the simple distance calculation formula calculates how much time must ascend to bypass the cube. The distance to be traveled, which in this case is the "height of the obstacle," is not directly given. However, the natural language prompt provides information that the "obstacle" in the map is not higher than 5 meters". Thus, using this information with the ascent speed, the agent performs the vertical maneuver for a fixed amount of time, effectively avoiding the obstacle. This solves the issue of providing the agent with 3D environment information, which can increase computation time.

Mission 3 replaces the cube obstacle with a spherical obstacle to observe avoidance by circumvention. The UAV is instructed to remain outside the "clearance boundary" of the sphere.

\begin{figure}[htbp]
\centering
\includegraphics[width=3.3in]{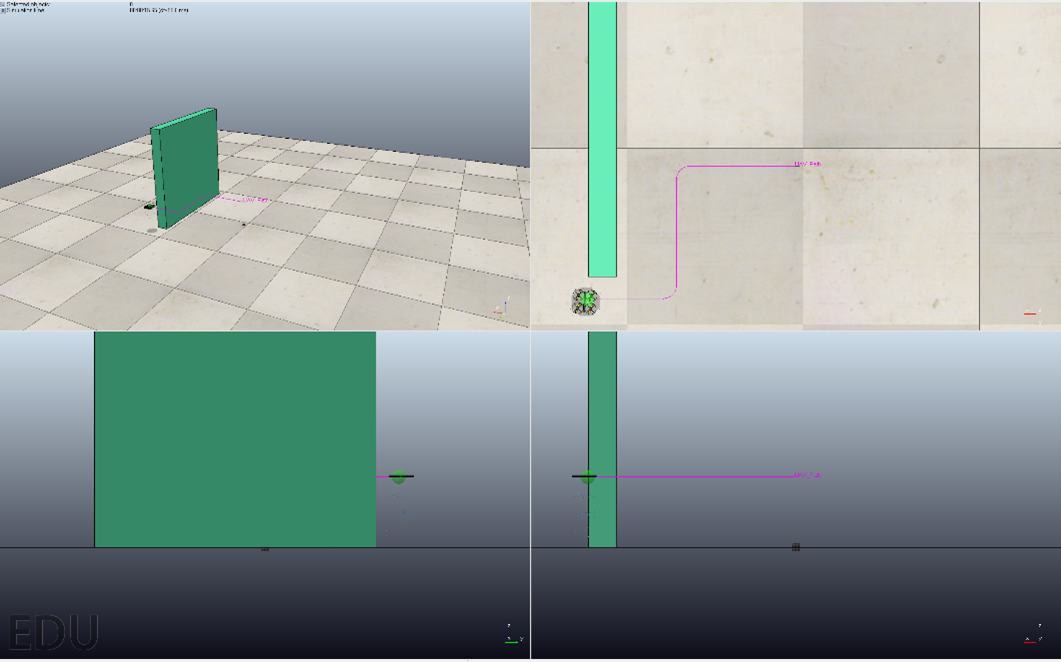}
\caption{Final execution of Mission 1 in CEP-2 environment (Direct LLM control)}
\label{fig11} 
\end{figure}

\begin{figure}[htbp]
\centering
\includegraphics[width=3.3in]{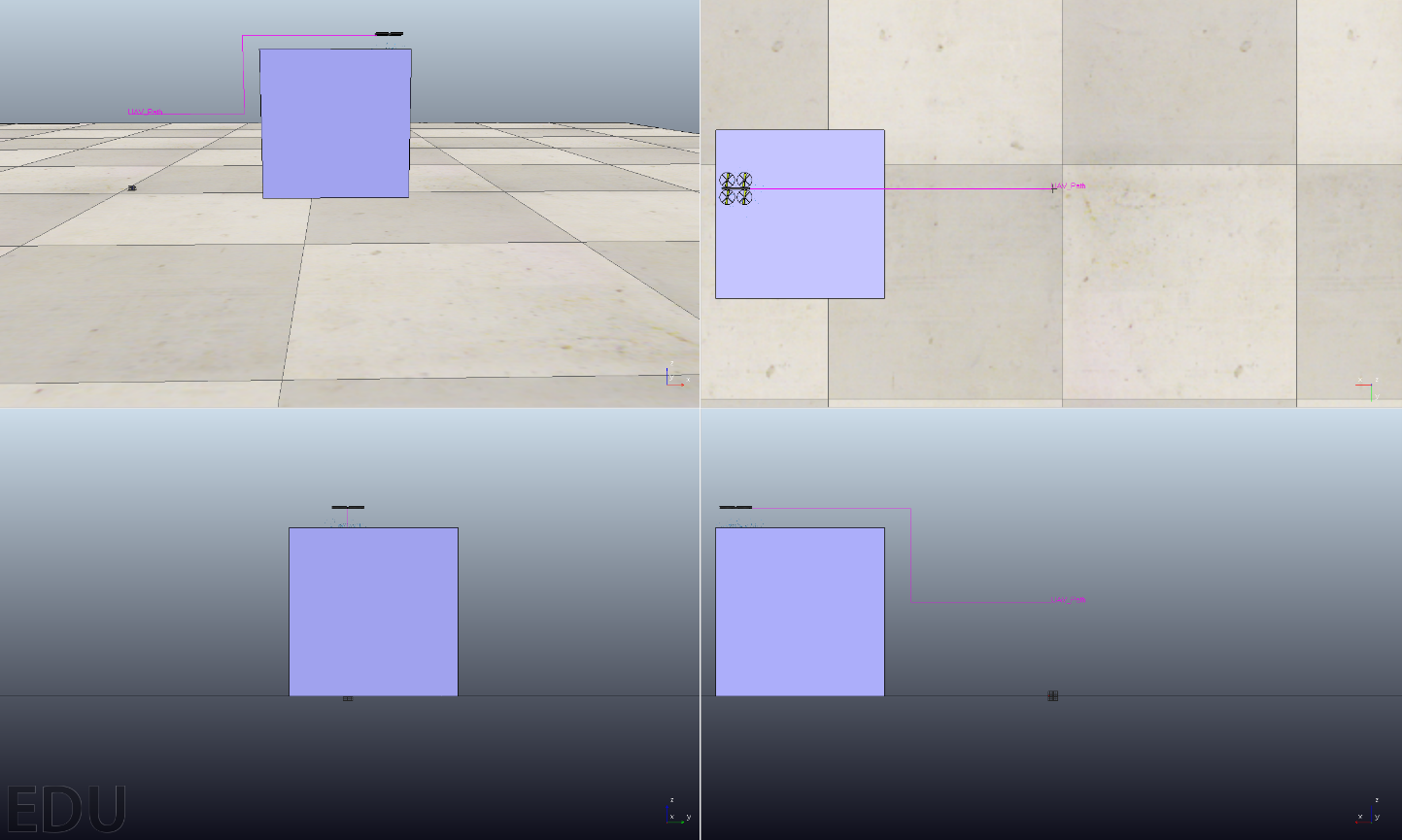}
\caption{Final execution of Mission 2 in CEP-2 environment (Direct LLM control)}
\label{fig12} 
\end{figure}

\begin{figure}[htbp]
\centering
\includegraphics[width=3.3in]{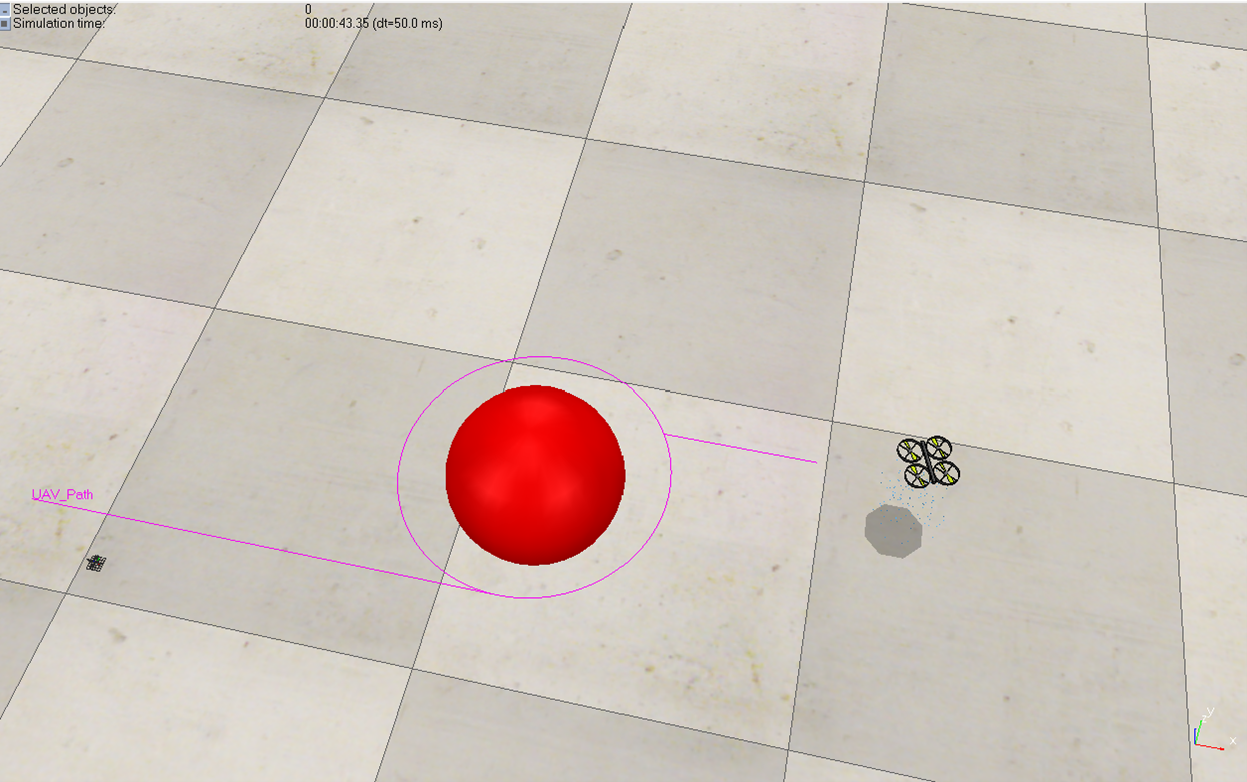}
\caption{Final execution of mission 3 in CEP-2 environment (Direct LLM control)}
\label{fig13} 
\end{figure}

Figures \ref{fig11}, \ref{fig12}, and \ref{fig13} show missions 1, 2, and 3 running on CEP-2 through the complete pathway shown in Figure \ref{fig7}.

\begin{figure*}[htbp]
\centering
\includegraphics[width=5.5in]{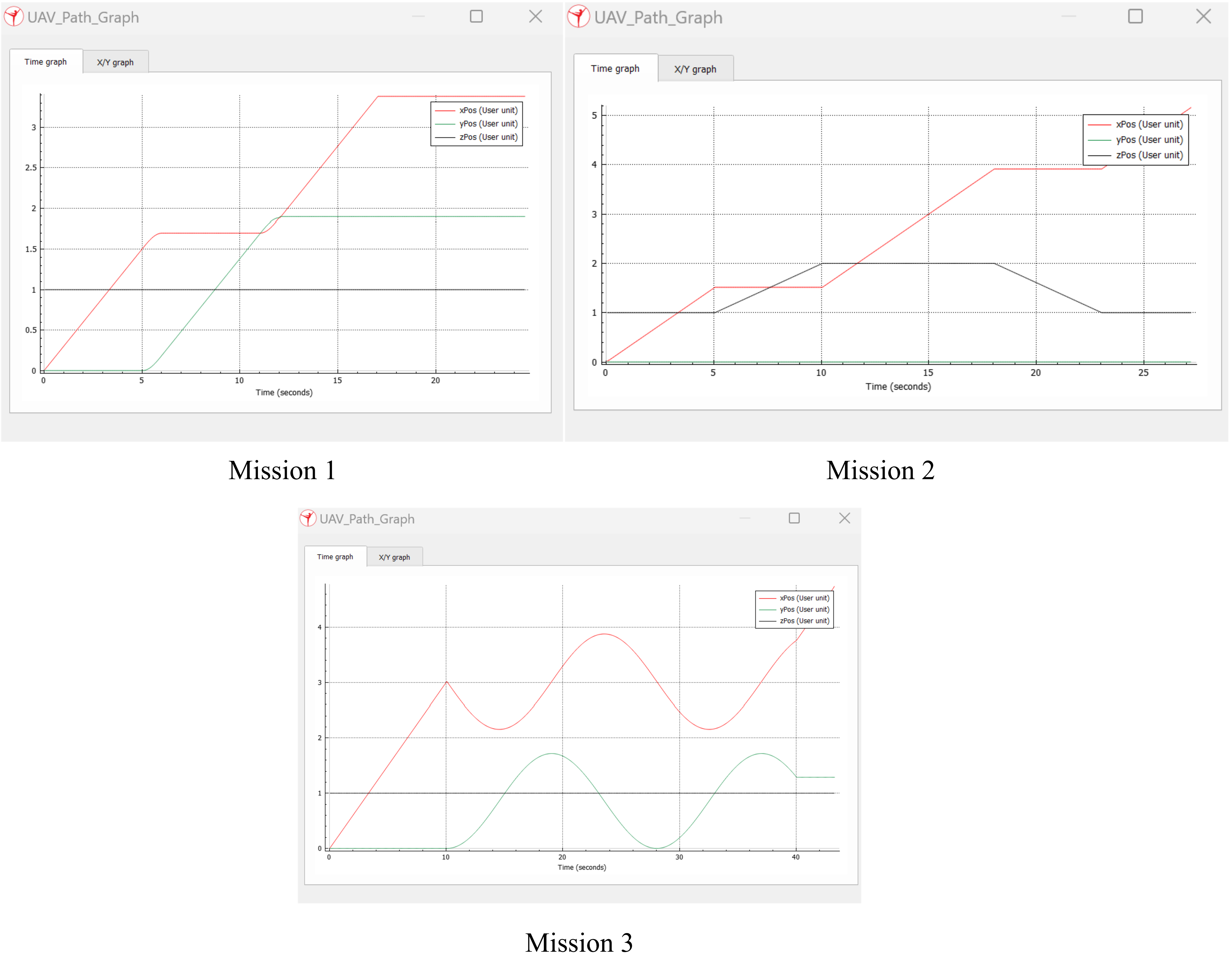}
\caption{Path changes for the three missions when executed under LLM control showing successful obstacle avoidance}
\label{fig14} 
\end{figure*}

Path changes for Missions 1, 2, and 3 in CEP-2 through LLM control are shown in Figure \ref{fig14}.

Currently, simple, controlled missions are expected to have $<$10 API calls, whereas complex missions in dynamic environments with additional sensors, interaction streams, and enabling agents with navigation, communication, and data collection capabilities are expected to require 100 calls or above. Advancements in LLM models may further reduce the I/O token and API calls, allowing extended complex missions.

\begin{figure}[htbp]
\centering
\includegraphics[width=3.4in]{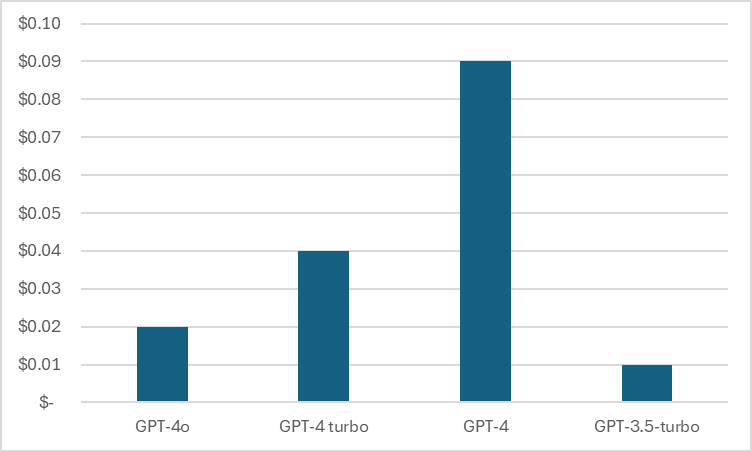}
\caption{API costs for various ChatGPT LLM models to be used in the experiment}
\label{fig15} 
\end{figure}

\section{Future Work and Conclusions}

The demonstrated concept effectively showed the integration of LLM wrappers such as MATGPT that harness existing LLM models and popular open-source robotic simulation platforms such as CoppeliaSim to produce effective and efficient interactions between robots and AI. By controlling the resolution of the produced interaction streams, achieving a finer level of control over the agent is possible. For example, the demonstrated interaction stream can be restructured and subdivided to incorporate additional functions when performing obstacle avoidance. Some initial suggestions are instructing the agent to capture obstacle information through onboard vision sensors. This is particularly useful in real-world situations where it can be necessary to re-examine sensor information if an agent collides with an obstacle \cite{Phadke2023Drone2DroneAS}. Other functions, such as referencing the obstacle's dimensions to consider the best possible modifications to the trajectory, are also necessary. For example, an obstacle that is wider but shorter in height can be bypassed more efficiently by an altitude modification to the trajectory waypoints rather than a horizontal deviation. A related advantage to this framework is that improvements to any component module are guaranteed to produce better results in the resultant interactions. Initial results show adequate control of the agent, with results as positive as zero net collisions between the agent and generated obstacles in a controlled environment.

The proposed concept and methodology have certain assumptions and limitations that warrant further study and development.
\begin{itemize}
  \item Obstacle presence is severely limited in the environment. As increased obstacle presence and avoidance activity can increase processing time and API calls, this reduced environment complexity is intended to keep initial projected costs low.
  \item Environment tests were constrained to simulation only, and environment information was pre-mapped and known to the UAV.
  \item The interaction speed response is slow. This is due to the multiple data and cross-function calls between the three platforms. Connectivity and hardware issues may also delay this speed, although a suitable benchmark evaluation was not used. However, as a result, the simulation runs on a delayed response clock wherein the entire experiment timeline is adjusted to account for data and information transfer between the simulation platform, the control section, and the LLM response time. While advancements in hardware, LLMs, and simulation platforms will significantly reduce this time requirement, a practical evaluation and performance benchmark is required to understand how these results will differ from real-world tests.
\end{itemize} 

The use of LLMs in swarm control is also a relatively less explored area. Swarm systems magnify the problem of agent control and resilience in dynamic environments, and a deeper, leveraged integration of LLMs is necessary for complex multi-agent flights to be possible.

This study provides the basis and the proof of concept for the "Language Driven Flight" framework being developed. Initial survey and development results indicate that integrating LLMs in robotics (both industrial and vehicular) is an obvious and important step in developing autonomous, assistive, and resilient technology.

\end{document}